# Fast Image Processing with Fully-Convolutional Networks

Qifeng Chen[*]    Jia Xu[*]    Vladlen Koltun

Intel Labs

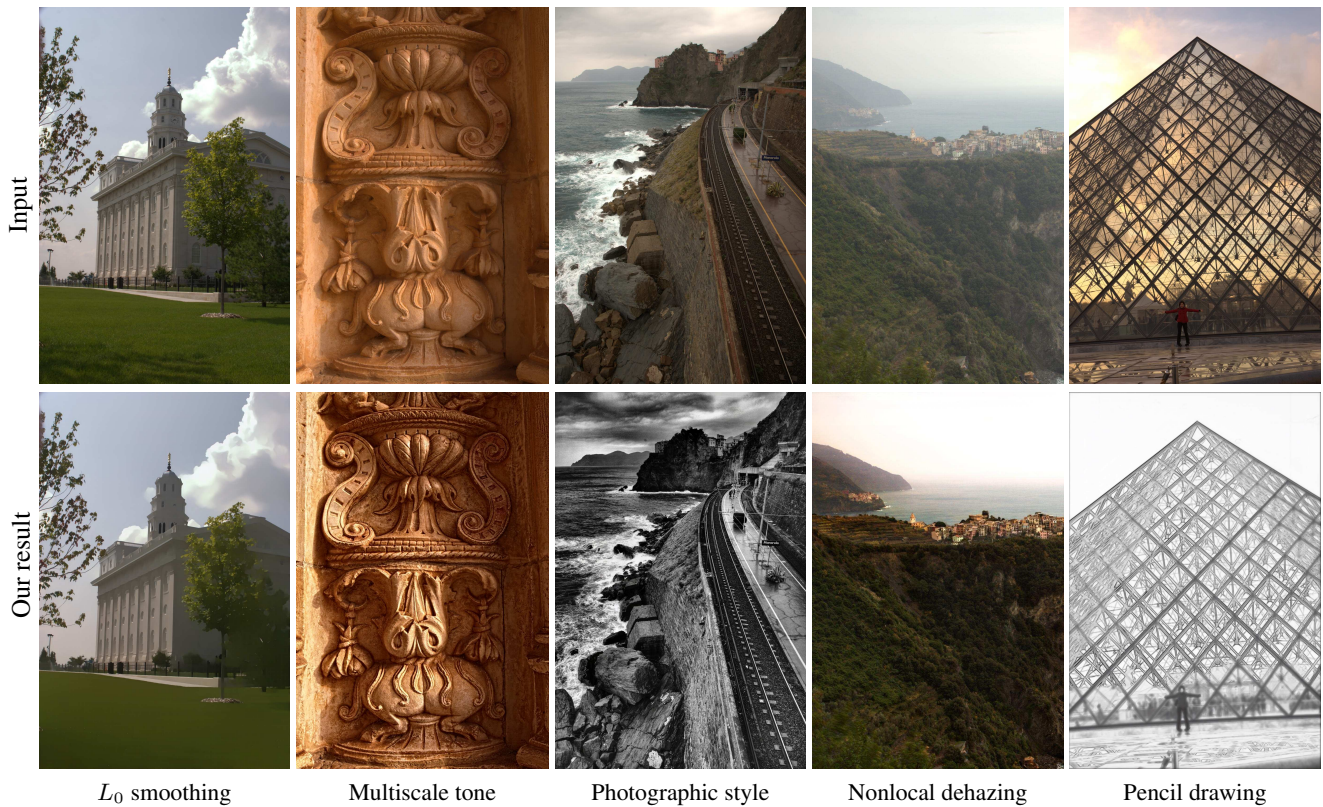

$L_0$ smoothing    Multiscale tone    Photographic style    Nonlocal dehazing    Pencil drawing

Figure 1. We present an approach to approximating image processing operators. This figure shows the results for five operators: $L_0$ gradient minimization, multiscale tone manipulation, photographic style transfer, nonlocal dehazing, and pencil drawing. All operators are approximated by the same model, with the same set of parameters and the same flow of computation.

## Abstract

*We present an approach to accelerating a wide variety of image processing operators. Our approach uses a fully-convolutional network that is trained on input-output pairs that demonstrate the operator's action. After training, the original operator need not be run at all. The trained network operates at full resolution and runs in constant time. We investigate the effect of network architecture on approximation accuracy, runtime, and memory footprint, and identify a specific architecture that balances these considerations. We evaluate the presented approach on ten advanced image processing operators, including multiple variational models, multiscale tone and detail manipulation, photographic style transfer, nonlocal dehazing, and nonphotorealistic stylization. All operators are approximated by the same model. Experiments demonstrate that the presented approach is significantly more accurate than prior approximation schemes. It increases approximation accuracy as measured by PSNR across the evaluated operators by 8.5 dB on the MIT-Adobe dataset (from 27.5 to 36 dB) and reduces DSSIM by a multiplicative factor of 3 compared to the most accurate prior approximation scheme, while being the fastest. We show that our models generalize across datasets and across resolutions, and investigate a number of extensions of the presented approach.*

---

[*]Joint first authors

# 1. Introduction

Research in image processing has yielded a variety of advanced operators that produce visually striking effects. Techniques developed in the last decade can dramatically enhance detail [24, 69, 26, 28, 60], transform the image by applying a master photographer's style [7, 5], smooth the image for the purpose of abstraction [73, 76, 79], and eliminate the effects of atmospheric scattering [25, 35, 27, 9]. This is accomplished by a variety of algorithmic approaches, including variational methods, gradient-domain processing, high-dimensional filtering, and manipulation of multiscale representations.

The computational demands and running times of existing operators vary greatly. Some operators, such as bilateral filtering, have benefitted from more than a decade of concerted investment in their acceleration. Others still take seconds or even minutes for high-resolution images. While most existing techniques can be accelerated by experts given sufficient research and development time, such acceleration schemes often require significant expertise and may not generalize across operators.

One general approach to accelerating a broad range of image processing operators is well-known: downsample the image, execute the operator at low resolution, and upsample [45, 34, 14]. This approach suffers from two significant drawbacks. First, the original operator must still be evaluated on a lower-resolution image. This can be a severe handicap because some operators are slow and existing implementations cannot be executed at interactive rates even at low resolution. Second, since the operator is never evaluated at the original resolution, its effects on the high-frequency content of the image may not be modeled properly. This can limit the accuracy of the approximation.

In this paper, we investigate an alternative approach to accelerating image processing operators. Like the downsample-evaluate-upsample approach, the presented method approximates the original operator. Unlike the downsampling approach, the method operates on full-resolution images, is trained end-to-end to maximize accuracy, and does not require running the original operator at all. To approximate the operator, we use a convolutional network that is trained on input-output pairs that demonstrate the action of the operator. After training, the network is used in place of the original operator, which need not be run at all.

We investigate the effects of different network architectures in terms of three properties that are important for accelerating image processing operators: approximation accuracy, runtime, and compactness. We identify a specific architecture that satisfies all three criteria and show that it approximates a wide variety of standard image processing operators extremely accurately. We evaluate the presented approach on ten advanced image processing operators, including multiple forms of variational image smoothing, adaptive detail enhancement, photographic style transfer, and dehazing. All operators are approximated using an identical architecture with no hyperparameter tuning. Five of the trained approximators are demonstrated in Figure 1, which shows their action on images from the MIT-Adobe 5K test set (not seen during training).

For all evaluated operators, the presented approximation scheme outperforms the downsampling approach. For example, the PSNR of our approximators across the ten considered operators on the MIT-Adobe test set is 36 dB, compared to 25 dB for the high-accuracy variant of bilateral guided upsampling [14]. At the same time, our approximators are faster than the fastest variant of that scheme. Our approximators run in constant time, independent of the runtime of the original operator.

We conduct extensive experiments that demonstrate that our simple approach outperforms a large number of recent and contemporary baselines, and that trained approximators generalize across datasets and to image resolutions not seen during training. We also investigate a number of extensions and show that the presented approach can be used to create parameterized networks that expose parameters that can be used to interactively control the effect of the image processing operator at test time; to train a single network that can emulate many diverse image processing operators and combine their effects; and to process video.

# 2. Related Work

Many schemes have been developed for accelerating image processing operators. The bilateral filter in particular has benefitted from long-term investment in its acceleration [21, 72, 15, 59, 2, 1, 29, 8]. Another family of dedicated acceleration schemes addresses the median filter and its variants [72, 61, 54, 80]. Other work has examined the acceleration of variational methods [6, 62, 13, 17], gradient-domain techniques [46], convolutions with large spatial support [23], and local Laplacian filters [5]. (Deep mathematical connections between these families of operators exist [57].) While many of these schemes successfully accelerate their intended families of operators, they do not have the generality we seek.

A general approach to accelerating image processing operators is to downsample the image, evaluate the operator at low resolution, and upsample [45, 34, 14]. This approach accelerates a broad range of operators by approximating them. It is largely agnostic to the operator but requires that the operator avoid spatial transformation so that the original image can be used to guide the upsampling. (E.g., no spatial warping such as perspective correction.) Our method shares a number of characteristics with the downsampling approach: it targets a broad range of operators, uses an approximation, and assumes that the spatial layout of the im-age is preserved. However, our approximation has a much richer parameterization that can model the operator's effect on the high-frequency content of the image. Once trained, the approximator does not need to execute the original operator at all. We will show that our method is more accurate than the downsampling approach on a wide range of tasks, while being faster.

Other work on accelerating image processing considers the system infrastructure and programming language. Given a powerful cloud backend and a bandwidth-limited network connection, high-resolution processing can be offloaded to the cloud [32]. Domain-specific languages can be used to schedule image processing pipelines to better utilize available hardware resources [63, 36]. Our work is complementary and provides an approach to approximating a wide variety of operators with a uniform parameterization. Such uniform parameterization and predictable flow of computation can assist further acceleration using dedicated hardware.

The closest works to ours are due to Xu et al. [75], Liu et al. [51], and Yan et al. [77]. We review each in turn. Xu et al. [75] used deep networks to approximate a variety of edge-preserving filters. Our work also uses deep networks, but differs in key technical decisions, leading to substantially broader scope and better performance. Specifically, the approach of Xu et al. operates in the gradient domain and requires reconstructing the output image by integrating the gradient field produced by the network. Since their networks produce non-integrable gradient fields, the authors had to constrain the final image reconstruction by introducing an additional data term that forces the output to be similar to the input. For this and other reasons, the approach of Xu et al. only applies to edge-preserving smoothing, has limited approximation accuracy, exhibits high running times (seconds for 1 MP images), and requires operator-specific hyperparameter tuning. In comparison, we train an approximator end-to-end, pixels to pixels, using a parameterization that is deeper and more context-aware while being more compact. We will demonstrate experimentally that the presented approach yields higher accuracy and lower runtimes while fitting a much bigger family of operators.

Liu et al. [51] combined a convolutional network and a set of recurrent networks to approximate a variety of image filters. This approach is quite flexible and outperforms the approach of Xu et al. on some operators, but does not achieve the approximation accuracy and speed we seek. We will show that a single convolutional network can achieve higher accuracy, while being faster and more compact.

Yan et al. [77] also applied deep networks to image adjustment. This work is also related to ours in its idea of approximating image transformations by deep networks. However, our work differs substantially in scope, technical approach, and results. Yan et al. use a fully-connected network that operates on each pixel separately. The network itself has a receptive field of a single pixel. Contextual information is only provided by hand-crafted input features, instead of being collected adaptively by the network. This places a substantial burden on manual feature design. In contrast, our approximator is a single convolutional network that is trained end-to-end, aggregates spatial context from the image as needed, and does not rely on extraneous modules or preprocessing. This leads to much greater generality, higher accuracy, and faster runtimes.

Deep networks have been used for denoising [39, 11, 3], super-resolution [10, 19, 40, 42, 41, 48, 50], deblurring [74], restoration of images corrupted by dirt or rain [22], example-based non-photorealistic stylization [30, 70, 40], joint image filtering [49], dehazing [64], and demosaicking [31]. None of the approaches described in these works were intended as broadly applicable replacements for the standard downsample-evaluate-upsample approach to image processing acceleration. Indeed, our experiments have shown that many approaches lack in either *speed*, *accuracy*, or *compactness* when applied across a broad range of operators. These criteria will be explored further in the next section.

## 3. Method

### 3.1. Preliminaries

Let $\mathbf{I}$ be an image, represented in the RGB color space. Let $f$ be an operator that transforms the content of an image without modifying its dimensions: that is, $\mathbf{I}$ and $f(\mathbf{I})$ have the same resolution. We will consider a variety of operators $f$ that use a broad range of algorithmic techniques. Our goal is to approximate $f$ with another operator $\hat{f}$, such that $\hat{f}(\mathbf{I}) \approx f(\mathbf{I})$ for all images $\mathbf{I}$. Note that the resolution of $\mathbf{I}$ is not restricted: both the operator $f$ and its approximation $\hat{f}$ are assumed to operate on variable-resolution images. Furthermore, we will consider many operators $\{f_i\}$ but require that our corresponding approximations $\{\hat{f}_i\}$ all share the same parameterization: same set of parameters, same flow of computation. The approximations will differ only in their parameters, which will be fit for each operator during training.

Our goal is to find a broadly applicable approach to accelerating image processing operators. We have identified three desirable criteria for such an approach. *Accuracy*: We seek an approach that provides high approximation accuracy across a broad range of popular image processing operators. *Speed*: The approach must be fast, ideally achieving interactive rates on HD images. *Compactness*: We seek an approach that can potentially be deployed within the constraints of mobile devices. An ideal network would have a very compact parameterization that can fit into on-chip SRAM, and a small memory footprint [33].

Our basic approach is to approximate the operator using a convolutional network [47]. The network must operate on variable-resolution images and must produce an output image at the same resolution as the input. This is known as dense prediction [52]. In principle, any fully-convolutional network architecture can be used for this purpose. Specifically, any network that has been used for a pixelwise classification problem such as semantic segmentation can instead be trained with a regression loss to produce continuous color rather than a discrete label per pixel. However, not all network architectures will yield high accuracy in this regime and most are not compact.

We have experimented with a large number of network architectures derived from prior work in high-level vision, specifically on semantic segmentation. We found that when some of these high-level networks are applied to low-level image processing problems, they generally outperform dedicated architectures previously designed for these image processing problems. The key advantage of architectures designed for high-level vision is their large receptive field. Many image processing operators are based on global optimization over the entire image, analysis of global image properties, or nonlocal information aggregation. To model such operators faithfully, the network must collect data from spatially distributed locations, aggregating information at multiple scales that are ultimately large enough to provide a global view of the image.

In Section 3.2 we describe an architecture that strikes the best balance between the different desiderata according to our experiments. Three alternative fully-convolutional architectures are described in the supplement.

## 3.2. Context aggregation networks

Our primary architecture is the multi-scale context aggregation network (CAN), developed in the context of semantic image analysis [78]. Its intermediate representations and its output have the same resolution as the input. Contextual information is gradually aggregated at increasingly larger scales, such that the computation of each output pixel takes into account all input pixels within a window of size exponential in the network's depth. This accomplishes global information aggregation for high-resolution images with a very compact parameterization. We will see that this architecture fulfills all of the desiderata outlined above.

We now describe the parameterization in detail. The data is laid out over multiple consecutive layers: $\{\mathbf{L}^0, \ldots, \mathbf{L}^d\}$. The first and last layers $\mathbf{L}^0, \mathbf{L}^d$ have dimensionality $m \times n \times 3$. These represent the input and output images. The resolution $m \times n$ varies and is not given in advance.

Each intermediate layer $\mathbf{L}^s$ ($1 \leq s \leq d-1$) has dimensionality $m \times n \times w$, where $w$ is the width of (i.e., the number of feature maps in) each layer. The content of intermediate layer $\mathbf{L}^s$ is computed from the content of the previous layer $\mathbf{L}^{s-1}$ as follows:

$$\mathbf{L}_i^s = \Phi\left(\Psi^s\left(b_i^s + \sum_j \mathbf{L}_j^{s-1} *_{r_s} \mathbf{K}_{i,j}^s\right)\right). \quad (1)$$

Here $\mathbf{L}_i^s$ is the $i^{\text{th}}$ feature map of layer $\mathbf{L}^s$, $\mathbf{L}_j^{s-1}$ is the $j^{\text{th}}$ feature map of layer $\mathbf{L}^{s-1}$, $b_i^s$ is a scalar bias, and $\mathbf{K}_{i,j}^s$ is a 3×3 convolution kernel. The operator $*_{r_s}$ is a dilated convolution with dilation $r_s$. The dilated convolution operator is the means by which the network aggregates long-range contextual information without losing resolution. Specifically, for image coordinates $\mathbf{x}$:

$$\left(\mathbf{L}_j^{s-1} *_{r_s} \mathbf{K}_{i,j}^s\right)(\mathbf{x}) = \sum_{\mathbf{a}+r_s\mathbf{b}=\mathbf{x}} \mathbf{L}_j^{s-1}(\mathbf{a}) \mathbf{K}_{i,j}^s(\mathbf{b}). \quad (2)$$

The effect of dilation is that the filter is tapped not at adjacent locations in the feature map, but at locations separated by the factor $r_s$. The dilation is increased exponentially with depth: $r_s = 2^{s-1}$ for $1 \leq s \leq d-2$. For $\mathbf{L}^{d-1}$, we do not use dilation. For the output layer $\mathbf{L}^d$ we use a linear transformation (1×1 convolution with no nonlinearity) that projects the final layer into the RGB color space.

For the pointwise nonlinearity $\Phi$, we use the leaky rectified linear unit (LReLU) [55]: $\Phi(x) = \max(\alpha x, x)$, where $\alpha = 0.2$. $\Psi^s$ is an adaptive normalization function, described in Section 3.3. Additional specification of the CAN architecture is provided in the supplement.

The network aggregates global context via full-resolution intermediate layers. It has a large receptive field while being extremely compact. It also has a small memory footprint during the forward pass. Since no skip connections across non-consecutive layers are employed, only two layers need to be kept in memory at any one time. Since the layers are all structurally identical, two fixed memory buffers are sufficient, with data flowing back and forth between them.

## 3.3. Adaptive normalization

We have found that using batch normalization improves approximation accuracy on challenging image processing operators such as style transfer and pencil drawing, but degrades performance on other image processing operators. We thus employ adaptive normalization that combines batch normalization and the identity mapping:

$$\Psi^s(x) = \lambda_s x + \mu_s BN(x), \quad (3)$$

where $\lambda_s, \mu_s \in \mathbb{R}$ are learned scalar weights and $BN$ is the batch normalization operator [37]. The weights $\{\lambda_s, \mu_s\}$ are learned by backpropagation alongside all other parameters of the network [67]. Learning these weights allows the model to adapt to the characteristics of the approximated operator, adjusting the strengths of the identity branch and the batch normalization branch as needed.

### 3.4. Training

The network is trained on a set of input-output pairs that contain images before and after the application of the original operator: $\mathcal{D} = \{\mathbf{I}_i, f(\mathbf{I}_i)\}$. The parameters of the network are the kernel weights $\mathcal{K} = \{\mathbf{K}_{i,j}^s\}_{s,i,j}$ and the biases $\mathcal{B} = \{b_i^s\}_{s,i}$. These parameters are optimized to fit the action of the operator $f$ across all images in the training set. We train with an image-space regression loss:

$$\ell(\mathcal{K}, \mathcal{B}) = \sum_i \frac{1}{N_i} \left\| \hat{f}(\mathbf{I}_i; \mathcal{K}, \mathcal{B}) - f(\mathbf{I}_i) \right\|^2, \qquad (4)$$

where $N_i$ is the number of pixels in image $\mathbf{I}_i$. This loss minimizes the mean-squared error (MSE) in the RGB color space across the training set. Although MSE is known to have limited correlation with perceptual image fidelity [71], experiments will demonstrate that training an approximator to minimize MSE will also yield high accuracy in terms of other measures such as PSNR and SSIM.

We have also experimented with more sophisticated losses, including perceptual losses that match feature activations in a visual perception network [10, 20, 40, 48, 16] and adversarial training [20, 38, 48]. We found that the higher-level feature matching losses did not increase approximation accuracy in our tasks; the image processing operators we target are not semantic in nature and can be approximated well by directly fitting the operator's action on the photographic content of the image. Adversarial training is known to be unstable [4, 56, 16] and we found that it also did not increase the already excellent results that we were able to obtain with an appropriate network architecture and a direct image-space loss.

Creating the training set $\mathcal{D}$ only requires running the original operator $f$ on a set of images. Training can thus be conducted on extremely large datasets that can be generated automatically without human intervention, although we found that training on a few thousand images already produces approximators that generalize well.

In order to expose the training to the effects of the operator $f$ on images of different resolutions, we use images of varying resolution for training. Specifically, given a set of high-resolution images, each is automatically resized to a random resolution between 320p and 1440p (e.g., 517p) while preserving its aspect ratio. These resized images are used for training. Training uses the Adam solver [43] and proceeds for 500K iterations (one randomly sampled image per iteration). This takes roughly one day on our test workstation.

## 4. Experiments

**Experimental setup.** We evaluate the presented approach on ten image processing operators: Rudin-Osher-Fatemi [66], TV-$L^1$ image restoration [58], $L_0$ smoothing [73], relative total variation [76], image enhancement by multiscale tone manipulation [24], multiscale detail manipulation based on local Laplacian filtering [5, 60], photographic style transfer from a reference image [5], dark-channel dehazing [35], nonlocal dehazing [9], and pencil drawing [53]. The operators, their effect on images, and our reference implementations are described in the supplement.

We use two image processing datasets: MIT-Adobe 5K and RAISE [12, 18]. MIT-Adobe 5K contains 5,000 high-resolution photographs covering a broad range of scenes, subjects, and lighting conditions. We use the default 2.5K/2.5K training/test split. The RAISE dataset contains 8,156 high-resolution RAW images captured by four photographers over a period of three years, depicting different scenes and moments across Europe. We use 2.5K randomly sampled images for training and 1K other randomly sampled images for testing.

We ran all ten operators on all images from the training and test sets of both datasets. For each operator, the input-output pairs from the MIT-Adobe training set were used for training. The same models and training procedures were used for all operators. The only difference between the ten approximators is in the output images that were provided in the training set. For each architecture, this procedure yielded ten identically parameterized models, trained to approximate the respective operators. These approximators are used for most of the experiments, which are conducted on the MIT-Adobe test set.

The same procedure was performed using the RAISE training set. This yielded models trained to approximate the same operators on the RAISE dataset. These models will be used to test cross-dataset generalization.

**Main results.** Our primary baseline is bilateral guided upsampling (BGU) [14], the state-of-the-art form of the downsample-evaluate-upsample scheme for accelerating image processing operators. There are two variants of the BGU approach, both with publicly available implementations. The first uses global optimization and is designed to approximate the original operator as closely as possible. The second is an approximation scheme designed to maximize speed, which was implemented in Halide [63] with specific attention to parallelization, vectorization, and data locality. We will compare to both variants of BGU, referred to respectively as BGU-opt and BGU-fast. We use the public implementations with the default parameters.

We also compare to a large number of baseline approaches that have used deep networks for related problems. The closest of these are the deep edge-aware filters of Xu et al. [75] and the recursive filters of Liu et al. [51]. Beyond this, we also evaluate the image transformation approach of Johnson et al. [40], which was developed for style transfer and superresolution but can be applied more broadly. Finally, we compare to the contemporaneous work of Isola et al. [38], who proposed an approach to "image-to-image translation" based on adversarial training. The approach of Isola et al. differs from the other baselines in that it is not fully-convolutional and is set up to operate at fixed resolution ($256 \times 256$). We report results for two versions of this baseline: one in which the output images are upsampled to the original resolution by bilinear interpolation, and one in which the output is upsampled using BGU-opt.

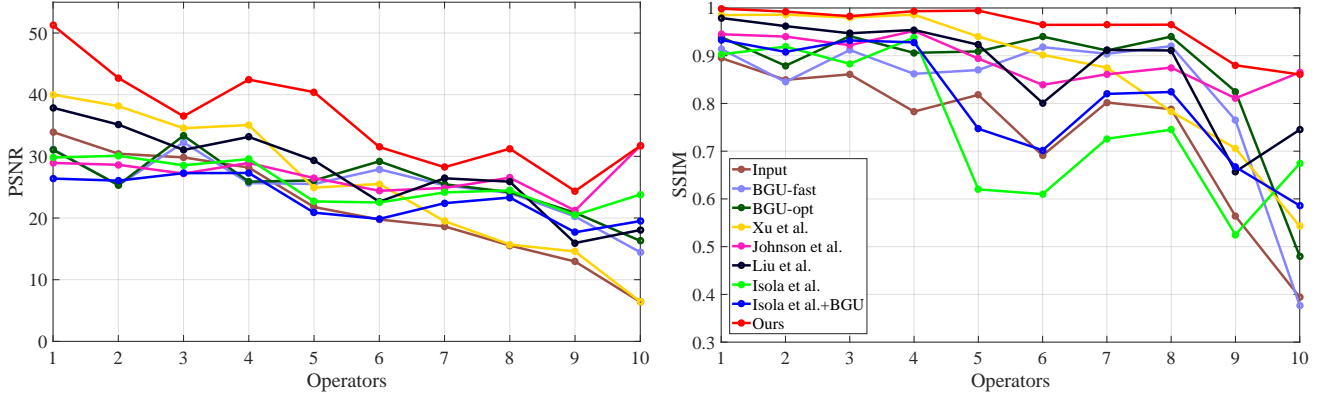

Figure 2. Approximation accuracy on the MIT-Adobe test set. Operators are arranged along the horizontal axis. From 1 to 10: Rudin-Osher-Fatemi [66], TV-$L^1$ image restoration [58], $L_0$ smoothing [73], relative total variation [76], image enhancement by multiscale tone manipulation [24], multiscale detail manipulation based on local Laplacian filtering [5, 60], nonlocal dehazing [9], dark-channel dehazing [35], photographic style transfer from a reference image [5], and pencil drawing [53].

Approximation accuracy achieved by each approach for each of the ten operators is visualized in Figure 2. All the numerical results are listed in the supplement. Our default model is a CAN with adaptive normalization, using $d = 9$ and $w = 24$ for the depth and width, respectively. This is the model referred to as 'Ours' in Figure 2 and Table 1. For each image, the output of each approach is compared to the output of the original reference operator, and the distance between the two images is evaluated in terms of PSNR and SSIM [71]. For each operator, the results are averaged over the MIT-Adobe test set. We also use a trivial baseline for calibration, referred to as Input. This trivial baseline simply uses the input image with no modification and thus evaluates the distance between the input image and the output of the reference operator. The Input baseline shows how a trivial approximation scheme (doing nothing) would fare and also provides an indication of how strongly the reference operator alters the image.

Due to the high computational demands of some of the reference operators, all images were scaled to 1080p resolution (∼1.75 MP) for this comprehensive experiment. We will evaluate cross-resolution performance in a subsequent experiment. Note that a resolution of 1080p had no special significance during training: the models were trained on images with randomly sampled resolution.

Average accuracy and runtime for each approach across all ten operators is summarized in Table 1. The runtime of each approach on each specific operator is reported in the supplement. The CAN parameterization is extremely compact: the network has a total of 37K parameters. It approximates the reference operators extremely accurately, achieving SSIM above 0.99 on four of the operators and SSIM above 0.96 on eight of them. (See the supplement for detailed results on the individual operators.)

| Method | MSE | PSNR | SSIM | Time (ms) | # of param |
|---|---|---|---|---|---|
| Reference | – | – | – | 9,502 | – |
| Input | 2607.9 | 21.75 | 0.745 | – | – |
| BGU-fast [14] | 521.8 | 24.70 | 0.827 | 320 | – |
| BGU-opt [14] | 413.5 | 25.27 | 0.865 | 2,378 | – |
| Xu et al. [75] | 2347.3 | 25.45 | 0.869 | 5,493 | 312K |
| Johnson et al. [40] | 215.0 | 26.89 | 0.890 | 203 | 1,678K |
| Liu et al. [51] | 383.8 | 27.56 | 0.879 | 458 | 152K |
| Isola et al. [38] | 279.5 | 25.62 | 0.754 | 198 | 57,184K |
| Isola et al. [38]+BGU | 457.2 | 23.07 | 0.805 | 2,352 | 57,184K |
| Ours | **59.1** | **36.04** | **0.960** | **190** | **37K** |

Table 1. Average accuracy, runtime, and number of parameters across all ten operators on the MIT-Adobe test set. Runtime is measured on images at 1080p resolution (∼1.75 MP).

Compared to our main baselines, BGU-opt and BGU-fast, our approach increases PNSR by 11 dB (from ∼25 to 36) and reduces DSSIM (=(1-SSIM)/2) by a multiplicative factor of 3. The downsampling approach does not perform well when the action of the operator at high resolution cannot be recovered from its output at low resolution. In contrast, our approach models the action of the operator directly at the original resolution. Our approach is also faster than BGU-fast and is more than an order of magnitude faster than BGU-opt. Runtime was measured on a workstation with an Intel i7-5960X 3.0GHz CPU and an Nvidia Titan X GPU. The runtime of BGU varies across operators, see the sup-plement for detailed results. The runtime of our approach is constant. It is 40 ms (25 fps) for 480p images, 190 ms for 1080p images, and scales linearly in the number of pixels. We used a standard deep learning library (TensorFlow) with no additional performance tuning.

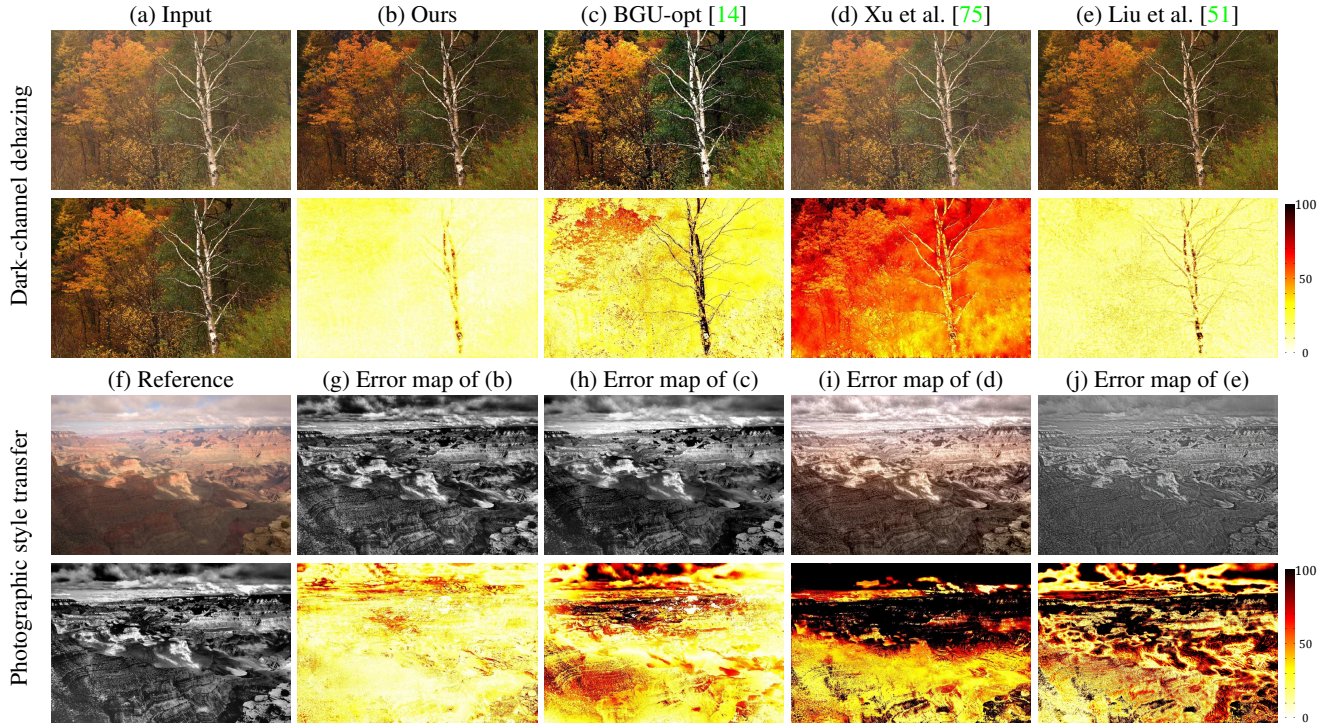

Figure 3. Qualitative results on images from the MIT-Adobe test set. For each operator, we show the input image, the result of the original reference operator, the result produced by our approximator, and results produced by BGU-opt [14], Xu et al. [75], and Liu et al. [51]. The error maps show per-pixel error, measured by Euclidean distance in 0-255 RGB space. Black indicates error of 100 or higher. Additional visualizations are provided in the supplement.

Of the prior approaches that use deep networks, Liu et al. [51] and Johnson et al. [40] achieve the best approximation accuracy. Our approach outperforms these baselines by 8.5 dB in PNSR, and reduces DSSIM by a multiplicative factor of 3. Our approach is also the fastest and has the most compact parameterization. Qualitative results are shown in Figure 3 and in the supplement.

**Additional experiments.** We now compare a number of different CAN configurations to alternative fully-convolutional architectures. These alternative architectures – Plain, Encoder-decoder [65], and FCN-8s [52, 68] – are described in detail in the supplement. All these models are trained by the same procedure as the CAN.

The results are summarized in Table 2. Here CAN24+AN is our primary model, referred to as 'Ours' in Table 1 ($d = 9$, $w = 24$, adaptive normalization). CAN32+AN is a more accurate but slower configuration ($d = 10$, $w = 32$, adaptive normalization). This configuration benefits from a receptive field of $513 \times 513$ versus the $257 \times 257$ receptive field of CAN24. We also evalu-ate two other variants of CAN32, controlling for the effect of adaptive normalization: CAN32 (no normalization) and CAN32+BN (BatchNorm). Finally, Table 2 also reports the performance of a single network (CAN32+AN) that represents all ten operators; this network is described in Section 5.

| Method | MSE | PSNR | SSIM | Time (ms) | # of param |
|---|---|---|---|---|---|
| FCN-8s | 344.1 | 26.36 | 0.808 | 150 | 30,510K |
| Encoder-decoder | 177.9 | 34.90 | 0.950 | 139 | 7,760K |
| Plain | 369.7 | 32.05 | 0.920 | **118** | 75K |
| CAN32 | 133.4 | 35.52 | 0.956 | 162 | 75K |
| CAN32+BN | 129.9 | 28.64 | 0.929 | 243 | 75K |
| CAN24+AN | 59.1 | 36.04 | 0.960 | 190 | **37K** |
| CAN32+AN | **36.0** | **37.59** | **0.966** | 277 | 75K |
| Single network | 110.3 | 29.86 | 0.931 | 385 | 78K |

Table 2. Average accuracy, running time, and number of parameters of different network architectures over all ten operators on the MIT-Adobe test set. Running time is measured on images at 1080p resolution (∼1.75 MP).

**Cross-resolution generalization.** We now test how the trained approximators generalize across resolutions. To keep the time of the experiment manageable, we focus on the $L_0$ smoothing operator for this purpose. Recall that our approximator was trained on images resized to random resolutions between 320p and 1440p. We now compare the trained model to baselines on a set of specific resolutions: 320p, 480p, 720p, 1080p, 1440p, and 2160p. For this purpose, the MIT-Adobe test set was resized to each of these resolutions, the reference operator was executed on these images, and all methods were evaluated at each resolution. The results are shown in the supplement. They indicate that the accuracy of our approximator is stable and outperforms the other approaches across resolutions. Note that the 2160p condition (∼7 MP) tests the generalization of our model to resolutions never seen during training. (The maximal resolution used during training was 1440p.)

**Cross-dataset generalization.** We have also evaluated how the trained operators generalize across datasets. To this end, for each operator, we tested two models on the MIT-Adobe test set: one trained on the MIT-Adobe training set and one trained on the RAISE training set. Similarly, for each operator, we tested two models on the RAISE test set: one trained on the RAISE training set and one trained on the MIT-Adobe training set. The detailed results are given in the supplement. They indicate that the trained approximators generalize extremely well and effectively represent the underlying action of the reference operators. The accuracy in corresponding conditions (e.g., MIT → MIT and RAISE → MIT) is virtually identical.

**Ablation studies.** Additional controlled experiments on network depth and width are reported in the supplement.

## 5. Extensions

We now describe three extensions of the presented approach: representing parameterized operators, representing multiple operators by a single network, and video processing.

**Parameterized operators.** An image processing operator can have parameters that control its action. For example, variational image smoothing operators [66, 58, 73] commonly have a parameter $\lambda$ that controls the relative strength of the regularizer: higher $\lambda$ leads to more aggressive smoothing. Other operators, such as multiscale tone manipulation, have multiple meaningful parameters that can be used to control the operator's effect [24]. Our approach extends naturally to creating parameterized approximators that expose these degrees of freedom at test time. To this end, we add channels to the input layer. For each parameter we wish to expose, we add an input channel that is used to communicate the parameter's value to the network. During training, we apply the operator with randomly sampled parameter values, thus showing the network the effect of the parameter on the operator. Quantitative results are reported in the supplement and qualitative results are shown in the video.

**One network to represent them all.** So far, we have trained separate networks for different operators, albeit with identical parameterizations. We now show that all 10 operators can be represented by a single network, which can emulate any of the individual operators at test time. This shows that a single compact network can execute a large number of advanced image processing effects at high accuracy. To this end, we augment the input layer by adding 10 additional channels, where each channel is a binary indicator that corresponds to one of the 10 operators. During training, we randomly sample an operator and an input image in each iteration. Training proceeds for 500K iterations total, as in the other experiments. For this experiment we use the CAN32 configuration with adaptive normalization.

The approximation accuracy achieved by the trained network across the 10 operators is reported in Table 2. The accuracy on each individual operator is given in the supplement. Remarkably, a single compact network that represents all 10 operators achieves high accuracy, well above the most accurate prior approximation scheme (compare to the results in Table 1). The trained network is demonstrated in the supplementary video. As shown in the video, the network can also smoothly transition between the operators when it is given continuous values in the auxiliary input channels, even though it was trained with one-hot vectors only.

**Video processing.** We also apply the trained models to videos from the Tanks and Temples dataset [44]. This further demonstrates cross-dataset generalization. (The models were trained on the MIT-Adobe dataset.) We simply apply the approximator to each frame. Although no provisions are made for temporal coherence, the results are as coherent as the original operators. The results are shown in the supplementary video.

## 6. Conclusion

We have presented an approach to approximating a wide range of image processing operators. All operators are approximated with the same parameterization and the same flow of computation. We have shown that the presented approach significantly outperforms prior approximation schemes.

We see the uniform and regular flow of computation in the presented model as a strong advantage. While the model is already faster than baselines using a generic implementation, we expect that significant further acceleration can be achieved.

## A. Operators

In this appendix, we describe in more detail the ten image processing operators used in our experiments. Our approach approximates all operators using the same model.

**Rudin-Osher-Fatemi.** Rudin-Osher-Fatemi (ROF) [18] is a seminal model for variational image restoration. The model aims to remove noise while preserving veridical image features by optimizing a variational objective over the image. Let $I : \Omega \to \mathbb{R}$ be a grayscale image. A restored image $J : \Omega \to \mathbb{R}$ can be computed by minimizing the following objective:

$$\int_\Omega |\nabla J| + \lambda \int_\Omega (I - J)^2, \quad (1)$$

where $\lambda$ is a free parameter that controls the smoothness of $J$. The first term $\int_\Omega |\nabla J|$ is the total variation regularization and the second term $\int_\Omega (I-J)^2$ is a data term that uses the $L^2$ norm. Equation (1) is strictly convex, so there is a unique global minimum.

**TV-$L^1$.** TV-$L^1$ [15] is a variational image restoration model that uses the following objective:

$$\int_\Omega |\nabla J| + \lambda \int_\Omega |I - J|. \quad (2)$$

Unlike the ROF model, TV-$L^1$ uses the more robust $L^1$ norm in the data term. Objective (2) is convex but not strictly convex, so the global minimizer may not be unique.

**$L_0$ smoothing.** The $L_0$ smoothing operator [21] makes use of the $L_0$ norm in the regularization term. This operator globally identifies the most important edges by penalizing the number of non-zero gradients in the image. The objective has the following form:

$$\int_\Omega |\nabla J|_0 + \lambda \int_\Omega (I - J)^2. \quad (3)$$

The objective is highly non-convex and cannot be optimized by traditional gradient-based methods. We use the solver provided by Xu et al. [21].

Objective (3) dates back to the work of Geman and Geman [5] and Mumford and Shah [13]. This objective is known as the Potts model or the piecewise-constant Mumford-Shah model. In addition to the solver of Xu et al. [21], which we use as the reference operator in our work, there are other recent solvers that optimize this objective [20, 14].

**Relative total variation.** Relative total variation (RTV) [23] is a model for extracting image structure by suppressing detail. This is also a variational model. It differs from the preceding ones by the form of the regularizer. The objective is

$$\int_\Omega \left( \frac{D_x}{L_x + \varepsilon} + \frac{D_y}{L_y + \varepsilon} \right) + \lambda \int_\Omega (I - J)^2, \quad (4)$$

where $D_x = G * |\partial_x J|$, $D_y = G * |\partial_y J|$, $L_x = |G * \partial_x J|$, $L_x = |G * \partial_x J|$, $G$ is a Gaussian kernel, and $\varepsilon$ is a small positive number. This objective is non-convex. We use the solver provided by Xu et al. [23].

**Multiscale tone manipulation.** This operator enhances an image by boosting features at multiple scales [4]. The method constructs a three-level image decomposition: a base layer $B$ and two detail layers, $D_1$ and $D_2$. The base layer is simply the LAB lightness channel of the input image $I$. The detail layers are constructed as $D_1 = B - \Psi(B)$ and $D_2 = \Psi(B) - \Psi(\Psi(B))$, where $\Psi(\cdot)$ denotes edge-preserving smoothing via weighted least-squares optimization. A new image can be constructed by nonlinearly combining these layers:

$$M + S\left(\delta_0(B - M)\right) + S(\delta_1 D_1) + S(\delta_2 D_2), \quad (5)$$

where $(\delta_0, \delta_1, \delta_2)$ are parameters, $M$ is a constant image with the mean intensity of $B$, and $S(\cdot)$ is a sigmoid function. Different sets of parameters boost features at different scales. We use the implementation of Farbman et al. [4] and use the default parameters to generate coarse-scale, medium-scale, and fine-scale images. These are then averaged to yield the final output.

**Detail manipulation.** This is another approach to multi-scale detail manipulation, based on local Laplacian filtering [16]. We use the accelerated implementation of Aubry et al. [1].

**Style transfer.** This operator transfers the photographic style of a reference image to the input image [1]. The operator is designed to transfer both local and global contrast and proceeds iteratively, alternating between local Laplacian filtering and histogram matching. We use the implementation of Aubry et al. [1] with their default style image.

**Dark-channel dehazing.** The goal of image dehazing is to remove some of the effects of atmospheric absorption and scattering. The standard image formation model used for this task is

$$I(\mathbf{x}) = t(\mathbf{x}) J(\mathbf{x}) + (1 - t(\mathbf{x})) A, \quad (6)$$

where $\mathbf{x}$ is a pixel, $I$ is the sensor irradiance, $J$ is the scene radiance, $A$ is the global atmospheric light, and $t$ is the transmission factor. Equation (6) is underconstrained and different dehazing techniques use different prior assumptions. Haze removal using the dark channel prior [7] is based on the observation that the atmospheric light can often be computed by identifying color channels that would have been dark in the absence of haze. We use the implementation of He et al. [7].

**Nonlocal dehazing.** This is a recent dehazing technique that uses a nonlocal prior [2]. It is based on the observation that pixel colors in haze-free images are clustered in color space, and that haze spreads these clusters into radial lines. The atmospheric light and transmission factors are recovered by identifying these lines in color space, and haze is removed using Equation (6). We use the implementation of Berman et al. [2].

**Pencil drawing.** This is a nonphotorealistic image stylization technique that aims to reproduce the appearance of a color pencil drawing while retaining the spatial structure of the image [12]. The technique computes a stroke layer from the gradient map and combines it with a tone layer, computed by a parametric model that represents tone distributions of pencil sketches. We use the implementation of Lu et al. [12].

## B. Context Aggregation Networks

Here we provide an illustration and a further specification of the context aggregation network (CAN), our primary architecture for approximating image processing operators. The context aggregation architecture is illustrated schematically in Figure 4. For the purpose of this figure, we use depth $d = 6$ and width $w = 8$. The dilation is increased from $r_1 = 1$ in $\mathbf{L}^1$ to $r_4 = 8$ in $\mathbf{L}^4$. The commensurate growth in the receptive field of each element in each layer can be seen in the figure. For $\mathbf{L}^{d-1}$ ($\mathbf{L}^5$ in Figure 4), we do not use dilation. For the output layer $\mathbf{L}^d$ ($\mathbf{L}^6$ in the figure) we use a linear transformation ($1 \times 1$ convolution with no nonlinearity) that projects the final feature layer into the RGB color space.

Figure 4 provides only a schematic visualization. The network we use is deeper and has a much larger receptive field. Table 3 provides a specification of the CAN32 configuration, which uses $d = 10$ and $w = 32$ and provides a receptive field of $513 \times 513$.

## C. Alternative Fully-Convolutional Architectures

In this appendix, we describe a number of fully-convolutional architectures that are evaluated alongside the CAN.

**Plain.** The first alternative architecture is a plain feedforward convolutional network that operates at full resolution. Specifically, we take the context aggregation network presented above and remove dilation. The network structure is the same, but all dilated convolutions are replaced by regular convolutions. The receptive field in the final layers of the network is $19 \times 19$. We use this architecture as a distinct baseline for two reasons. First, it isolates the effect of dilation (and therefore large receptive field) while retaining all the other desirable properties of the previously presented architecture. Second, it is analogous to an architecture that has recently been used for demosaicking and denoising [6], and thus illustrates the performance characteristics of this architecture when applied to a broad range of operators.

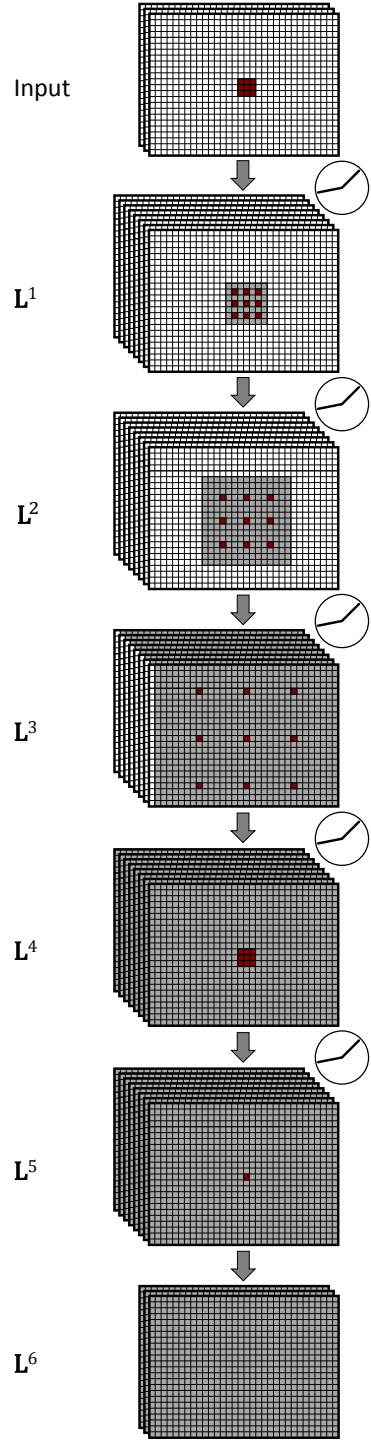

Figure 4: Schematic illustration of the context aggregation network. This visualization uses depth $d = 6$ and width $w = 8$. The red pixels show the application of dilated convolutions [24]. The shaded gray pixels show the receptive field of a single element. Circles show the nonlinear transformation $\Phi$. The model we use in practice is deeper and wider than shown here.

| Layer | 1 | 2 | 3 | 4 | 5 | 6 | 7 | 8 | 9 | 10 |
|---|---|---|---|---|---|---|---|---|---|---|
| Convolution | 3×3 | 3×3 | 3×3 | 3×3 | 3×3 | 3×3 | 3×3 | 3×3 | 3×3 | 1×1 |
| Dilation | 1 | 2 | 4 | 8 | 16 | 32 | 64 | 128 | 1 | 1 |
| Receptive field | 3×3 | 7×7 | 15×15 | 31×31 | 63×63 | 127×127 | 255×255 | 511×511 | 513×513 | 513×513 |
| Nonlinearity | ✓ | ✓ | ✓ | ✓ | ✓ | ✓ | ✓ | ✓ | ✓ | ✗ |
| Width | 32 | 32 | 32 | 32 | 32 | 32 | 32 | 32 | 32 | 3 |

Table 3: A specification of the CAN32 configuration.

**Encoder-decoder.** The next architecture highlights an alternative way to achieve a large receptive field: progressively reducing the resolution of the feature layers and then increasing them back to the original resolution. Such hourglass-shaped networks are sometimes referred to as encoder-decoders. Of course, the high-frequency content that is lost in the internal layers due to downsampling must be recovered somehow. A standard solution is to add skip connections across non-consecutive layers, for example connecting all layers that have the same resolution on the two sides of the hourglass. Our reference encoder-decoder architecture is the u-net [17]. The network has 23 convolutional layers. Each encoding layer applies 3×3 convolutions, followed by truncation, max pooling, and downsampling. With each downsampling step, the number of feature channels is doubled. The decoder performs upsampling by 2×2 upconvolutions, concatenates the result with the corresponding feature maps from one of the encoding layers, and applies 3×3 convolutions and truncations. The final layer applies a 1×1 convolution that projects each feature column into the RGB color space.

We make two modifications to the original u-net architecture [17]. First, to reduce computation time and memory footprint, we use half of the filters in each layer (e.g., 32 rather than 64 in the first layer): we found that this is sufficient to get high accuracy and it matches our configuration of the other baselines. Second, we pad each layer when necessary to make the output image match the size of the input. This makes our implementation agnostic to aspect ratio, whereas the original u-net requires the image to be square.

We will see that the encoder-decoder achieves comparable accuracy to the context aggregation network across operators. Furthermore, due to the low resolution of the intermediate layers, it is even faster. However, due to the high width of the intermediate layers, its capacity (number of parameters), is two orders of magnitude higher: roughly 7.7 million as opposed to 75 thousand for CAN32. Furthermore, due to the skip connections across the network, up to half of the layers must be kept in memory during the forward pass, increasing the network's memory footprint.

**FCN-8s.** As a reference baseline, we also use the fully-convolutional setup of the VGG-16 network (specifically, FCN-8s) [11, 19]. This network also performs downsampling and then upsampling, although asymmetrically: most of the capacity is in the downsampling layers. This network is fast, but is far from compact: more than 30 million parameters. The biggest issue, however, is that its approximation accuracy is low, due to the severe internal downsampling and limited support for recovering the lost high-frequency content during upsampling. This will be illustrated in the experiments.

## D. Accuracy and Runtime

Here we provide the complete quantitative results for the approximation accuracy and runtime of different approaches. The approximation accuracy of each approach on each operator is given in Table 4. These are the numerical results that are visualized in Figure 2 and summarized in Table 1 in the paper. The approximation accuracy for different CAN configurations and alternative fully-convolutional architectures is given in Table 5; these are the numerical results that are summarized in Table 2 in the paper.

The running time of each approach on each operator is given in Table 6. The operators are arranged in the same order as in Table 4. Runtime was measured on a workstation with an Intel i7-5960X 3.0GHz CPU and an Nvidia Titan X GPU. Our approach is faster than BGU-opt by more than an order of magnitude. It is faster than BGU-fast on eight of the ten operators.

## E. Cross-Resolution Generalization

The cross-resolution generalization results on $L_0$ smoothing are shown in Figure 5.

## F. Cross-Dataset Generalization

Here we provide the precise results of the cross-dataset generalization experiment. For each operator, we tested two models on the MIT-Adobe test set: one trained on the MIT-Adobe training set and one trained on the RAISE training set. Similarly, for each operator, we tested two models on

| Method | Rudin-Osher-Fatemi [18] | | | TV-$L^1$ [15] | | | $L_0$ smoothing [21] | | | Relative total variation [23] | | | Multiscale tone [4] | | |
|---|---|---|---|---|---|---|---|---|---|---|---|---|---|---|---|
| | MSE | PSNR | SSIM | MSE | PSNR | SSIM | MSE | PSNR | SSIM | MSE | PSNR | SSIM | MSE | PSNR | SSIM |
| Input | 31.4 | 33.94 | 0.895 | 91.9 | 30.43 | 0.850 | 69.1 | 29.83 | 0.861 | 120.5 | 28.18 | 0.783 | 464.6 | 21.79 | 0.818 |
| BGU-fast [3] | 56.7 | 31.17 | 0.914 | 252.0 | 25.31 | 0.845 | 42.6 | 32.26 | 0.912 | 205.9 | 25.68 | 0.862 | 193.1 | 25.52 | 0.870 |
| BGU-opt [3] | 59.8 | 31.06 | 0.939 | 254.6 | 25.35 | 0.879 | 34.3 | 33.36 | 0.941 | 197.7 | 25.90 | 0.906 | 168.0 | 26.09 | 0.909 |
| Xu et al. [22] | 8.8 | 40.02 | 0.985 | 14.3 | 38.15 | 0.986 | 24.3 | 34.57 | 0.980 | 22.6 | 35.08 | 0.986 | 225.6 | 24.92 | 0.940 |
| Johnson et al. [9] | 139.6 | 28.93 | 0.945 | 146.8 | 28.63 | 0.940 | 174.8 | 27.21 | 0.922 | 140.5 | 28.93 | 0.952 | 211.1 | 26.49 | 0.894 |
| Liu et al. [10] | 13.1 | 37.88 | 0.979 | 25.4 | 35.16 | 0.962 | 52.6 | 31.04 | 0.947 | 35.8 | 33.16 | 0.954 | 83.1 | 29.35 | 0.923 |
| Isola et al. [8] | 93.6 | 29.82 | 0.903 | 83.4 | 30.09 | 0.919 | 119.7 | 28.55 | 0.883 | 96.1 | 29.56 | 0.938 | 448.3 | 22.70 | 0.620 |
| Isola et al. [8]+BGU | 159.0 | 26.40 | 0.933 | 175.1 | 26.06 | 0.908 | 129.8 | 27.28 | 0.932 | 136.8 | 27.31 | 0.928 | 555.6 | 20.87 | 0.747 |
| Ours | **0.6** | **51.24** | **0.999** | **4.3** | **42.72** | **0.992** | **14.9** | **36.50** | **0.983** | **4.4** | **42.45** | **0.993** | **6.3** | **40.42** | **0.995** |
| Method | Detail manipulation [16] | | | Nonlocal dehazing [2] | | | Dark-channel dehazing [7] | | | Style transfer [1] | | | Pencil drawing [12] | | |
| | MSE | PSNR | SSIM | MSE | PSNR | SSIM | MSE | PSNR | SSIM | MSE | PSNR | SSIM | MSE | PSNR | SSIM |
| Input | 712.5 | 19.76 | 0.691 | 1048.2 | 18.65 | 0.802 | 2642.2 | 15.55 | 0.788 | 3762.9 | 12.95 | 0.564 | 17135.6 | 6.40 | 0.394 |
| BGU-fast [3] | 113.5 | 27.90 | 0.918 | 251.0 | 25.29 | 0.904 | 354.4 | 24.06 | 0.920 | 672.5 | 20.25 | 0.765 | 2521.8 | 14.45 | 0.377 |
| BGU-opt [3] | 82.7 | 29.19 | 0.940 | 247.9 | 25.47 | 0.911 | 345.3 | 24.20 | 0.940 | 582.4 | 20.86 | 0.825 | 1590.9 | 16.30 | 0.480 |
| Xu et al. [22] | 190.7 | 25.53 | 0.902 | 1001.7 | 19.50 | 0.875 | 2551.4 | 15.70 | 0.783 | 2701.8 | 14.58 | 0.706 | 16731.5 | 6.45 | 0.543 |
| Johnson et al. [9] | 263.1 | 24.42 | 0.839 | 301.9 | 24.88 | 0.861 | 204.2 | 26.53 | 0.875 | 521.6 | 21.21 | 0.811 | 46.7 | **31.70** | **0.865** |
| Liu et al. [10] | 366.2 | 22.65 | 0.801 | 180.4 | 26.44 | 0.912 | 221.3 | 25.90 | 0.911 | 1732.2 | 15.95 | 0.657 | 1127.9 | 18.06 | 0.745 |
| Isola et al. [8] | 445.3 | 22.52 | 0.610 | 306.6 | 24.15 | 0.726 | 282.4 | 24.48 | 0.745 | 627.5 | 20.47 | 0.525 | 291.5 | 23.81 | 0.674 |
| Isola et al. [8]+BGU | 697.7 | 19.86 | 0.702 | 398.6 | 22.41 | 0.820 | 337.8 | 23.32 | 0.824 | 1211.5 | 17.69 | 0.667 | 770.3 | 19.51 | 0.586 |
| Ours | **48.0** | **31.52** | **0.965** | **133.2** | **28.25** | **0.965** | **74.6** | **31.24** | **0.965** | **258.4** | **24.37** | **0.880** | **45.7** | **31.70** | 0.861 |

Table 4: Approximation accuracy on the MIT-Adobe test set.

| Method | Rudin-Osher-Fatemi [18] | | | TV-$L^1$ [15] | | | $L_0$ smoothing [21] | | | Relative total variation [23] | | | Multiscale tone [4] | | |
|---|---|---|---|---|---|---|---|---|---|---|---|---|---|---|---|
| | MSE | PSNR | SSIM | MSE | PSNR | SSIM | MSE | PSNR | SSIM | MSE | PSNR | SSIM | MSE | PSNR | SSIM |
| FCN-8s | 54.6 | 32.39 | 0.946 | 38.8 | 33.42 | 0.958 | 89.1 | 30.05 | 0.918 | 40.2 | 33.65 | 0.972 | 357.9 | 24.30 | 0.714 |
| Encoder-decoder | **0.5** | **51.96** | 0.999 | 2.3 | 45.22 | **0.995** | 12.6 | 37.27 | **0.992** | 3.6 | 43.06 | **0.997** | **3.1** | **43.81** | **0.997** |
| Plain | 0.7 | 50.06 | **1.000** | **1.9** | **46.91** | **0.995** | 27.9 | 33.90 | 0.980 | 9.2 | 38.85 | 0.983 | 12.8 | 37.66 | 0.989 |
| CAN32 | 0.6 | 51.29 | 0.999 | 3.7 | 43.44 | 0.993 | 12.6 | 37.32 | 0.988 | 3.8 | 42.99 | 0.994 | 3.8 | 42.57 | **0.997** |
| CAN32+BN | 69.2 | 31.30 | 0.963 | 86.3 | 30.22 | 0.950 | 136.5 | 27.81 | 0.927 | 95.6 | 29.76 | 0.955 | 128.0 | 28.12 | 0.953 |
| CAN24+AN | 0.6 | 51.24 | 0.999 | 4.3 | 42.72 | 0.992 | 14.9 | 36.50 | 0.983 | 4.4 | 42.45 | 0.993 | 6.3 | 40.42 | 0.995 |
| CAN32+AN | 0.6 | 51.86 | 0.999 | 3.0 | 44.57 | 0.993 | **12.1** | **37.48** | 0.987 | **3.3** | **43.55** | 0.995 | 4.9 | 41.60 | 0.996 |
| CAN32+AN+Single | 20.9 | 35.16 | 0.978 | 26.0 | 34.28 | 0.965 | 50.4 | 31.20 | 0.948 | 34.4 | 33.05 | 0.958 | 51.0 | 31.24 | 0.967 |
| Method | Detail manipulation [16] | | | Nonlocal dehazing [2] | | | Dark-channel dehazing [7] | | | Style transfer [1] | | | Pencil drawing [12] | | |
| | MSE | PSNR | SSIM | MSE | PSNR | SSIM | MSE | PSNR | SSIM | MSE | PSNR | SSIM | MSE | PSNR | SSIM |
| FCN-8s | 413.7 | 23.00 | 0.689 | 294.3 | 24.53 | 0.788 | 296.9 | 24.51 | 0.806 | 754.3 | 19.64 | 0.556 | 1101.6 | 18.14 | 0.735 |
| Encoder-decoder | 83.2 | 29.26 | 0.949 | 101.2 | 30.11 | **0.970** | 142.3 | 28.37 | 0.966 | 593.4 | 20.66 | 0.827 | 836.4 | 19.33 | 0.810 |
| Plain | 313.3 | 23.37 | 0.863 | 135.7 | 28.38 | 0.952 | 186.9 | 27.67 | 0.957 | 1712.0 | 15.99 | 0.689 | 1296.9 | 17.68 | 0.788 |
| CAN32 | **19.7** | **35.39** | **0.981** | 102.5 | 30.09 | 0.969 | 154.7 | 28.31 | 0.965 | 298.1 | 23.81 | 0.854 | 577.3 | 21.02 | 0.826 |
| CAN32+BN | 116.4 | 28.19 | 0.934 | 281.7 | 25.35 | 0.918 | 192.1 | 27.07 | 0.923 | 146.1 | 27.08 | 0.900 | 47.4 | 31.55 | 0.865 |
| CAN24+AN | 48.0 | 31.52 | 0.965 | 133.2 | 28.25 | 0.965 | 74.6 | 31.24 | 0.965 | 258.4 | 24.37 | 0.880 | 45.7 | 31.70 | 0.861 |
| CAN32+AN | 29.3 | 33.66 | 0.966 | **84.2** | **30.46** | **0.970** | **53.0** | **32.76** | **0.974** | **129.8** | **27.62** | **0.913** | **39.6** | **32.31** | **0.869** |
| CAN32+AN+Single | 72.7 | 29.67 | 0.938 | 137.2 | 27.99 | 0.951 | 212.7 | 26.06 | 0.932 | 345.7 | 23.27 | 0.850 | 152.3 | 26.72 | 0.825 |

Table 5: Approximation accuracy of different network architectures on the MIT-Adobe test set.

the RAISE test set: one trained on the RAISE training set and one trained on the MIT-Adobe training set. The results for all operators are shown in Table 7. They indicate that the trained approximators generalize extremely well. The accuracy in corresponding conditions (e.g., MIT → MIT and RAISE → MIT) is virtually identical. On the MIT test set, the SSIM achieved by models trained on RAISE is within 1% of the SSIM achieved by models trained on the MIT training set, for all operators. The same is true on the RAISE test set. This indicates that our approximators represent the underlying action of the reference operators effectively.

# G. Ablation Studies

Here we report the results of additional controlled experiments that study different aspects of our model's structure and their effect on approximation accuracy. For these experiments, we again use the $L_0$ smoothing operator on the MIT-Adobe dataset.

| Method | ROF [18] | TV-$L^1$ [15] | $L_0$ [21] | RTV [23] | Tone [4] | Detail [16] | Dehaze (NL) [2] | Dehaze (DC) [7] | Style [1] | Pencil [12] |
|---|---|---|---|---|---|---|---|---|---|---|
| Reference | 18,598 | 22,181 | 7,053 | 10,411 | 10,268 | 1,190 | 6,114 | 7,983 | 6,271 | 4,947 |
| BGU-fast [3] | 382 | 458 | 187 | 323 | 217 | 186 | 232 | 227 | 739 | 251 |
| BGU-opt [3] | 2,436 | 2,472 | 2,321 | 2,377 | 2,271 | 2,240 | 2,286 | 2,281 | 2,793 | 2,305 |
| Xu et al. [22] | 5,493 | 5,493 | 5,493 | 5,493 | 5,493 | 5,493 | 5,493 | 5,493 | 5,493 | 5,493 |
| Johnson et al. [9] | 203 | 203 | 203 | 203 | 203 | 203 | 203 | 203 | 203 | 203 |
| Liu et al. [10] | 458 | 458 | 458 | 458 | 458 | 458 | 458 | 458 | 458 | 458 |
| Isola et al. [8] | 198 | 198 | 198 | 198 | 198 | 198 | 198 | 198 | 198 | 198 |
| Isola et al. [8]+BGU | 2,352 | 2,352 | 2,352 | 2,352 | 2,352 | 2,352 | 2,352 | 2,352 | 2,352 | 2,352 |
| Ours | **190** | **190** | **190** | **190** | **190** | **190** | **190** | **190** | **190** | **190** |

Table 6: Running time (in milliseconds) on MIT-Adobe test set images at 1080p resolution ($\sim$1.75 MP).

| Train $\to$ Test | Rudin-Osher-Fatemi [18] | | | TV-$L^1$ [15] | | | $L_0$ smoothing [21] | | | Relative total variation [23] | | | Multiscale tone [4] | | |
|---|---|---|---|---|---|---|---|---|---|---|---|---|---|---|---|
| | MSE | PSNR | SSIM | MSE | PSNR | SSIM | MSE | PSNR | SSIM | MSE | PSNR | SSIM | MSE | PSNR | SSIM |
| MIT $\to$ MIT | 0.6 | 51.24 | 0.999 | 4.3 | 42.72 | 0.992 | 14.9 | 36.50 | 0.983 | 4.4 | 42.45 | 0.993 | 6.3 | 40.42 | 0.995 |
| RAISE $\to$ MIT | 2.5 | 46.52 | 0.996 | 3.5 | 43.96 | 0.993 | 17.2 | 35.96 | 0.984 | 4.6 | 42.38 | 0.993 | 8.5 | 39.46 | 0.993 |
| RAISE $\to$ RAISE | 1.8 | 46.09 | 0.996 | 4.2 | 43.07 | 0.991 | 15.7 | 36.31 | 0.983 | 5.0 | 42.14 | 0.992 | 7.7 | 39.58 | 0.994 |
| MIT $\to$ RAISE | 1.0 | 49.98 | 0.998 | 5.1 | 41.78 | 0.990 | 15.1 | 36.49 | 0.981 | 5.1 | 41.97 | 0.992 | 6.8 | 40.03 | 0.995 |
| | Detail manipulation [16] | | | Nonlocal dehazing [2] | | | Dark-channel dehazing [7] | | | Style transfer [1] | | | Pencil drawing [12] | | |
| | MSE | PSNR | SSIM | MSE | PSNR | SSIM | MSE | PSNR | SSIM | MSE | PSNR | SSIM | MSE | PSNR | SSIM |
| MIT $\to$ MIT | 48.0 | 31.52 | 0.965 | 133.2 | 28.25 | 0.965 | 74.6 | 31.24 | 0.965 | 258.4 | 24.37 | 0.880 | 45.7 | 31.70 | 0.861 |
| RAISE $\to$ MIT | 46.3 | 31.70 | 0.968 | 106.5 | 29.37 | 0.959 | 85.8 | 30.78 | 0.962 | 272.4 | 24.22 | 0.876 | 44.7 | 31.81 | 0.866 |
| RAISE $\to$ RAISE | 43.5 | 31.93 | 0.976 | 91.8 | 29.88 | 0.965 | 56.0 | 31.81 | 0.969 | 248.0 | 24.60 | 0.894 | 45.2 | 31.73 | 0.870 |
| MIT $\to$ RAISE | 48.7 | 31.44 | 0.972 | 138.8 | 27.89 | 0.967 | 72.2 | 30.87 | 0.968 | 265.7 | 24.34 | 0.889 | 49.8 | 31.38 | 0.863 |

Table 7: Cross-dataset generalization. Models were trained separately on the MIT-Adobe training set and the RAISE training set. Each model was then tested on the MIT-Adobe test set and the RAISE test set. The approximation accuracy is virtually identical in corresponding conditions ($< 1\%$ difference in SSIM), indicating that the trained approximators generalize well across datasets.

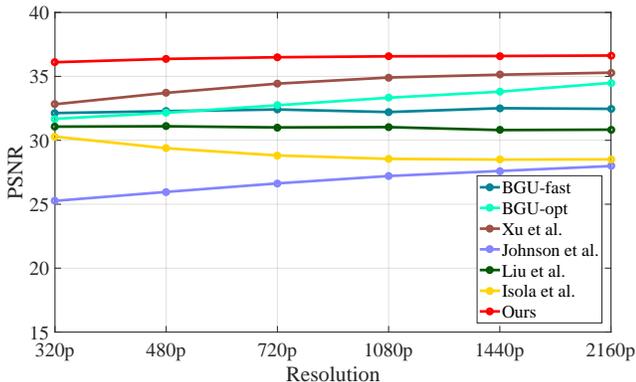

Figure 5: Cross-resolution generalization. Different approximation schemes were tested on images from the MIT-Adobe test set resampled to different resolutions. Our approach uses the same model for all resolutions. It outperforms the other approaches across resolutions, including on resolutions never seen during training.

**Depth.** We begin by training and testing the context aggregation network with different depths $d$. The results are given in Table 8. Note that smaller depth implies a smaller receptive field. As shown in the table, the results are good even for shallow networks: for example, the model achieves higher SSIM than BGU-opt even with depth 4. (With this depth, the running time on 1080p images is 67 ms.) The accuracy further improves with depth and saturates at $d = 9$.

| Depth | MSE | PSNR | SSIM | Time (ms) |
|---|---|---|---|---|
| 4 | 32.6 | 33.14 | 0.964 | **67** |
| 5 | 27.6 | 33.86 | 0.974 | 92 |
| 6 | 23.4 | 34.57 | 0.980 | 118 |
| 7 | 20.0 | 35.23 | 0.982 | 142 |
| 8 | 17.1 | 35.91 | 0.982 | 165 |
| 9 | **14.9** | **36.50** | 0.983 | 190 |
| 10 | 15.5 | 36.36 | **0.983** | 218 |
| 11 | 16.4 | 36.10 | **0.983** | 243 |

Table 8: Controlled evaluation of approximation accuracy as a function of depth $d$, with $w = 24$.

**Width.** We now evaluate the effect of width (the number of feature maps in each intermediate layer) on approximation accuracy. The experimental setup is the same as in the previous experiment. The results are reported in Table 9. The accuracy is again good even with a network that has fairly low capacity (8 feature maps per layer, 84 ms runtime at 1080p). Accuracy further increases with width.

| Width | MSE | PSNR | SSIM | Time (ms) |
|---|---|---|---|---|
| 8 | 32.3 | 33.12 | 0.954 | **84** |
| 16 | 19.7 | 35.30 | 0.979 | 131 |
| 24 | 14.9 | 36.50 | 0.983 | 190 |
| 32 | 13.5 | 36.93 | 0.986 | 249 |
| 48 | 12.6 | 37.26 | 0.988 | 388 |
| 64 | **9.9** | **38.30** | **0.989** | 517 |

Table 9: Evaluation of approximation accuracy as a function of width $w$, with $d = 9$.

## H. Parameterized Operators

Here we report in more detail the results on representing parameterized operators. We use the $L_0$ smoothing operator. We sample different hyperparamters $\lambda$ in Equation 3: $\lambda = \bar{\lambda} \exp(x)$ where $x$ is a random variable with uniform distribution $U(-\ln(10), \ln(10))$ and $\bar{\lambda} = 0.01$ is the default value, so $\lambda \in [0.1\bar{\lambda}, 10\bar{\lambda}]$. We train and test the approximator with randomly sampled parameters $\lambda$. The approximation accuracy achieved by our approach is 21.0 in MSE, 36.2 in PSNR, and 0.984 in SSIM.

## I. Qualitative Results

Extensive qualitative results are provided in a separate supplement. Our method consistently outperforms the other approaches. The most sophisticated prior downsample-evaluate-upsample scheme, BGU-opt, does not perform well when the action of the operator at high resolution cannot be recovered from its output at low resolution. In contrast, our method operates directly at the original resolution. Our direct approach is also more accurate than prior approaches that use deep networks.